\documentclass{sig-alternate-05-2015}

\usepackage{icvgip}
\usepackage[numbers]{natbib}
\usepackage{graphicx}
\usepackage{subfig}
\usepackage{multicol}
\usepackage{booktabs}
\usepackage{array}
\usepackage{multirow}
\usepackage{placeins}
\usepackage[pagebackref=true,letterpaper=true,colorlinks=true,bookmarks=false,urlcolor=blue]{hyperref}
\icvgipfinalcopy
\newcommand\myfigure{%
\resizebox{\textwidth}{!}{%
\begin{tabular}{c}
\includegraphics[height=4cm]{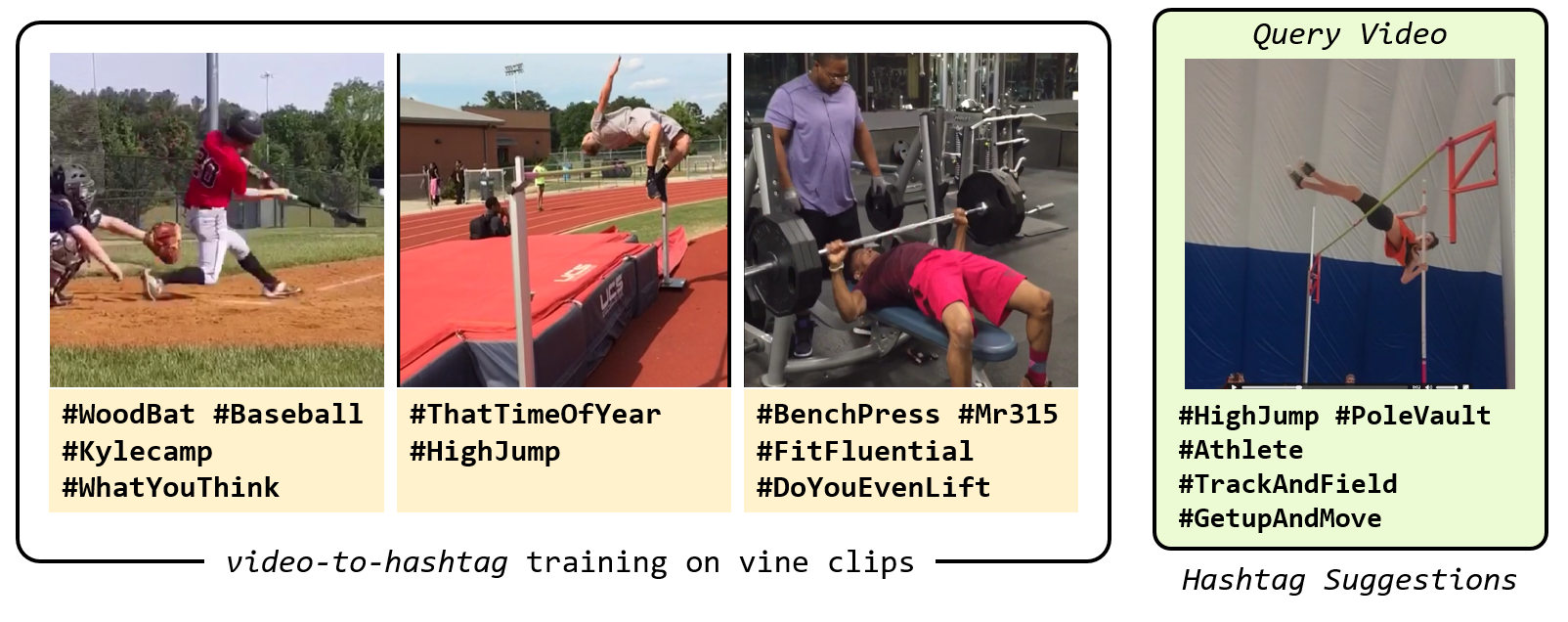}
\end{tabular}
}
\\[1.5\normalbaselineskip]
\vspace{2mm}
\refstepcounter{figure}\small{\noindent\parbox[l]{0.92\textwidth}{Figure~\thefigure. 
Learning a direct mapping from videos to hash-tags :  
sample frames from short video clips with user-given hash-tags (left); 
a sample frame from a query video and hash-tags suggested by our system for this query (right). 
}
}
\label{fig:vinetagsnadshots}
}

\begin{document}

\setcopyright{acmcopyright}
\conferenceinfo{ICVGIP}{'16 Guwahati, Assam India}
\title{Learning to Hash-tag Videos with Tag2Vec}
\author{Aditya Singh \hspace{1cm}  Saurabh Saini \hspace{1cm}  Rajvi Shah \hspace{1cm} P J Narayanan\\[0.5\normalbaselineskip]
\affaddr{CVIT, KCIS, IIIT Hyderabad, India\titlenote{\scriptsize{Center for Visual Information Technology (CVIT), 
Kohli Center on Intelligent Systems (KCIS), International Institute of Information Technology (IIIT) Hyderabad}}}\\[0.5\normalbaselineskip]
{\tt\small \{(aditya.singh,saurabh.saini,rajvi.shah)@research.,pjn@\}iiit.ac.in}\\[0.5\normalbaselineskip]
\small{\url{http://cvit.iiit.ac.in/research/projects/tag2vec}}
\vspace{-5mm}}
\CopyrightYear{2016}
\setcopyright{acmcopyright}
\conferenceinfo{ICVGIP,}{December 18-22, 2016, Guwahati, India}
\isbn{978-1-4503-4753-2/16/12}\acmPrice{\$15.00}
\doi{http://dx.doi.org/10.1145/3009977.3010035}
\maketitle

\begin{abstract}
User-given tags or labels are valuable resources 
for semantic understanding of visual media such 
as images and videos. 
Recently, a new type of labeling mechanism 
known as hash-tags have become increasingly 
popular on social media sites. In this paper, 
we study the problem of generating relevant 
and useful hash-tags for short video clips. 
Traditional data-driven approaches for 
tag enrichment and recommendation use 
direct visual similarity for label 
transfer and propagation. We attempt to 
learn a direct low-cost mapping from 
video to hash-tags using a two step training process. 
We first employ a natural language processing (NLP) technique,  
skip-gram models with neural network training to 
learn a low-dimensional vector 
representation of hash-tags (\emph{Tag2Vec}) 
using a corpus of $\sim$ 10 million hash-tags. 
We then train an embedding function 
to map video features to the low-dimensional Tag2vec 
space. We learn this embedding for 
29 categories of short video clips with hash-tags. 
A query video without any tag-information can then be 
directly mapped to the vector space of tags using the learned embedding 
and relevant tags can be found by performing a simple 
nearest-neighbor retrieval in the Tag2Vec space. 
We validate the relevance of the tags suggested by our system 
qualitatively and quantitatively with a user study.

\end{abstract}



\section{Introduction}
Over the last decade, social media websites such as Twitter, Instagram, Vine, YouTube have become increasingly popular.
These media sites allow users to upload, tag, and share their content with a wide audience across the world. 
In case of visual media such as images and videos, the user-given tags often provide rich semantic information about
the visual context as well as affective appeal of the media, otherwise hard to recognize and categorize.
Most popular image and video search engines still heavily rely upon user-given tags for relevant retrieval.
Apart from search and retrieval, hash-tags also facilitate browsing and content management. Varied web content can be 
organized easily using hash-tags which defines new trending concepts. From user perspective this makes it easier to share content 
and follow trends.

With increasing success of object category recognition algorithms on large data such as ImageNet,
automatic image tagging and captioning is showing remarkable progress. However, for videos with dynamic events, 
fewer attempts have been made at unsupervised video tagging. 

The methods for video annotation and tagging can be classified into two categories, (i) model 
based methods, and (ii) data-driven methods. Model based methods mainly apply several concept classifiers, 
pre-trained with low-level video features, and use the resulting concept labels for effective tagging. 
The shortcoming of this approach is revealed by the fact that it is impossible to enumerate all possible concept 
categories and their inter-relationships as perceptually understood. Data-driven approaches on the other hand do not explicitly 
discover or recognize visual concepts. They, instead directly propagate or transfer them
from tagged videos to query videos using some measure of visual similarity. 

In this paper, we present a hybrid approach for tag suggestion. We do not explicitly 
use pre-trained concept classifiers, nor do we use video-to-video similarity based measures. 
We propose a method to directly embed video features into a low-dimensional vector space of tag 
distribution. Given a query video, the relevant tags can than be retrieved by a simple nearest 
neighbour strategy. The \emph{video-to-tag} training is carried out in two stages. First, we 
learn a 100-dimensional vector space representation of popular tag words using a corpus of $\sim$ 10 million  
hash-tags using the algorithm of \citet{word2vec}. This algorithm trains a two-layer 
neural network with skip-gram representation to learn word embeddings in vector space. 
Extending the terminology of \cite{word2vec}, we call this vector space \emph{Tag2Vec} space in this paper. 
Second, we learn a nonlinear embedding of high-dimensional video features to the low-dimensional Tag2Vec 
space using a separate neural network \citet{zslearn}. For this task, we use $\sim$ 2740 short video 
clips from 29 categories and their associated hash-tags. Once trained, the final video-to-tag embedding 
can be leveraged to suggest tag words for query videos. Our approach is pictorially summarized 
in \autoref{fig:flowdiagram}. We evaluate the performance of our system 
qualitatively and quantitatively with a user study and show the method is promising. We also 
discuss limitations and future directions to improve effectiveness of this simple approach. 

The main contributions of this paper are as follows, (i) We study the problem of hash-tag suggestion for 
videos from a novel perspective and present a mechanism for direct embedding of videos to a 
vector space of hash-tags; (ii) We present a new dataset consisting of 3000 
random wild short social video clips spanning 29 categories with associated user-given 
hash-tags.

\section{Related Work}
In this section we present a brief discussion of the relevant literature related to our problem.
First, we discuss the methods related to the two core sub-systems of our method, tagging and word embedding, separately.
Later, we discuss the recent works on visual semantic joint understanding that leverage similar methodology. 

\subsection{Tagging}
Image and video tagging approaches can be coarsely categorized as model-based or data-driven.
\citet{correlativeMultilabel} combine the strengths of the two separate paradigms using separate binary classifiers 
learning and concept fusion. They focus on multi-label results and use gibbs random field based mathematical model. 
\citet{statModel4vAnR} construct a joint probability of visual region-based 
words with text annotations, incorporating co-occurrent visual features, and co-occurrent annotations to demonstrate 
that statistical methods can be used to retrieve videos by content. However, the main drawback of model-based approaches is the limit on 
detectors that can be trained. As there are thousands of concepts for which it is difficult to gather 
large training data for reliable learning. Due to this, there has been a shift from from model-based 
methods to data-driven similarity based methods. 

Data-driven methods utilize the abundance of videos shared by users and transfer tags based on 
similarity measures. \citet{TagSugLoc} proposed a video retagging approach based on visual and 
semantic consistency. This approach however only acknowledges tags which are nouns in the WordNet 
lexicon. Often the hash-tags are informal internet slang-words which rarely occur in proper language documents 
and hence do not have semantic consistency. \citet{SSLvideoConcept} use a graph 
based semi-supervised learning approach for manifold ranking. They use partial differential based 
anisotropic diffusion for label propagation. \citet{nearDuplicateVA} focuses on fast near duplicate 
video retrieval for automatic video annotation. They find near duplicates by indexing local features, 
fast pruning of false matches at frame levels, and localization of near
duplicate segments at video levels. Then a weighted majority approach is used for tag recommendation. 
\citet{graphRMvA}  devices a graph reinforcement framework to propagate tags developed by crawling
tags of similar videos for annotation by using text and visual features. \citet{BDMneighSim} computes similarity 
between two samples along with the difference in their
surrounding neighbourhood sample \& label distribution. The neighbourhood sample similarity is 
computed using KL divergence and label similarity is based on difference of label histograms of the two samples.
\citet{vtUsrSearchBeh} utilize the user click-through data along with the similarity based measures to tackle the problem of video tagging. 
Our approach is also based on data-driven similarity but instead of directly measuring video similarity, we learn a 
direct mapping to embed the videos in a lower-dimensional Tag2Vec space then use a nearest-neighbour classifier for tag suggestion.

\subsection{Word Embedding}
Common approaches for word embedding is through neural networks \cite{zeroShotLearning}, dimensionality reduction of co-occurrence matrix.
\cite{WE_hellingerPCA, neuralWE_iMatFact, weRevisited_eMatFact}, explicitly constructed probabilistic models \cite{eucEmb_coOucData} etc. 
In our method we use neural network based embedding as will be focusing only on such approaches here. 
Amongst the early approaches, \citet{neuralProbLangModel} reduces the high dimensionality of words representations in 
contexts by learning a distributed representation for words. They use a feed forward neural network with 
a linear projection layer and a non-linear hidden layer which jointly learns a word vector representation 
and a statistical language model. Currently, widely popular technique of \citet{word2vec} proposes method for 
learning word vectors from a large amount of unstructured data. They also show that the learned space is 
a metric space and meaningful algebraic operations can be performed on the word vectors.
\citet{bayesianNWE} proposes a bayesian skip-gram method which maps words to densities in a latent space rather 
than word vectors which results in less effort in hyperparameter tuning. 

\subsection{Visual-semantic Joint Embedding}
Vector representation of words is used by many recent approaches for joint visual semantic learning 
\cite{zeroShotLearning, gcpraction, showNtell_NImgCapGen, fastZeroShotImgTag, karpathy2015deep}.
\citet{zeroShotLearning} learns an embedding function which performs a mapping of image features to semantic word space.
Then they utilize this for categorization of seen and unseen classes. 
\citet{fastZeroShotImgTag} utilize linear mappings and non-linear neural networks to tag an image.
They define the problem of assigning tags as identification of a principal direction for an image in word space.
This principal direction ranks relevant tags ahead of irrelevant tags. Similar to all these approaches we also used a neural network 
for learning embedding function but we focus only on hash tags.
\cite{gcpraction} maps the video representations to semantic space for improving action classification. 
Recent image captioning methods \cite{showNtell_NImgCapGen, karpathy2015deep} have shown remarkable 
progress in generating rich descriptive sentences for natural images. These methods use 
recurrent neural network architectures with large data for training. 
Visual question answering (VQA) systems \cite{vqa} also employ deep learning to 
train an answering system for natural language queries.  Though deep features have shown 
promise in image based captioning and VQA systems, one either needs a huge amount of data 
to train deep networks or needs to fine-tune a pre-trained network. 
For videos such pre-trained networks are not readily available yet and  
though we have collected a dataset of nearly 3000 short videos, it is 
not sufficient to train a deep network. Hence, we use the state-of-the-art 
hand-crafted features for our application. 
On a related note, recently \citet{MovieQA} released an interesting video based 
question answering dataset by aligning book descriptions to movie scenes. 
Our work however has a different focus in its application to hash-tags and 
wild social video clips. 

The rest of the paper is organized as follows: \autoref{sec:method} explains the 
methodology and generation of tag space. \autoref{sec:exp} provides the experimental details, 
results and analysis. Finally in \autoref{sec:conclusion} we conclude our work and 
present future research directions.  

\section{Method}
\label{sec:method}

\begin{figure}
 \includegraphics[width=0.23\textwidth]{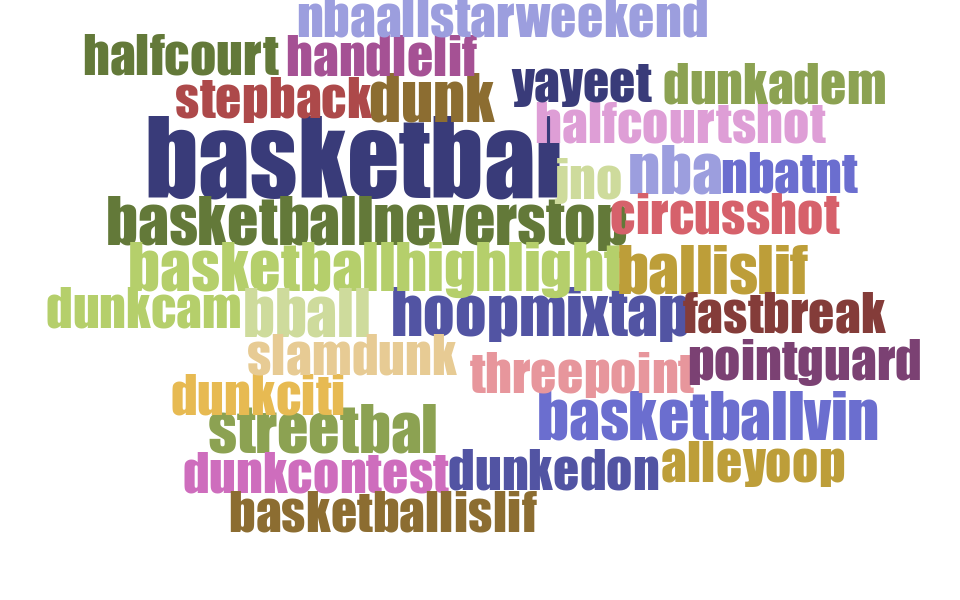}
 \includegraphics[width=0.23\textwidth]{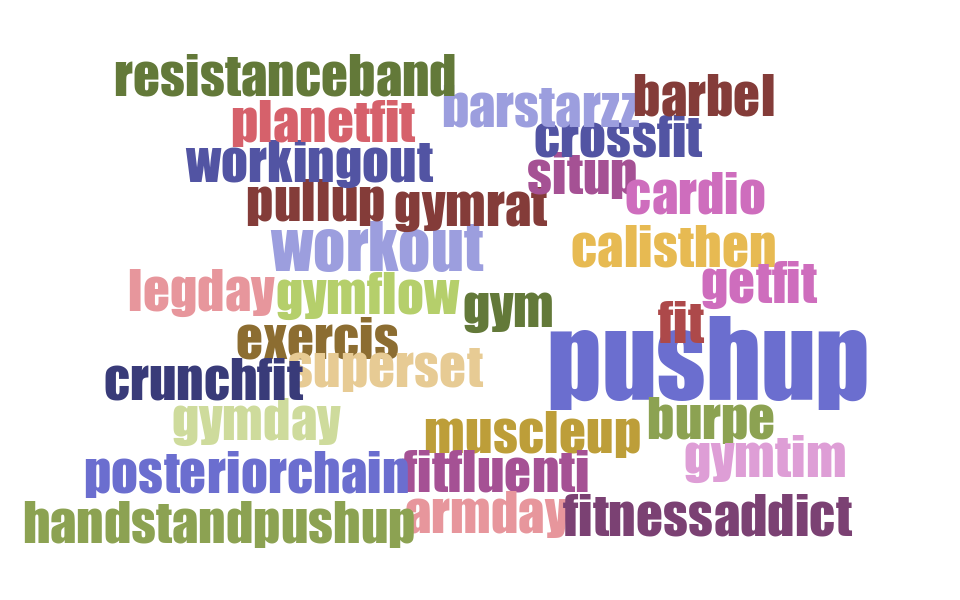}\\
 \includegraphics[width=0.23\textwidth]{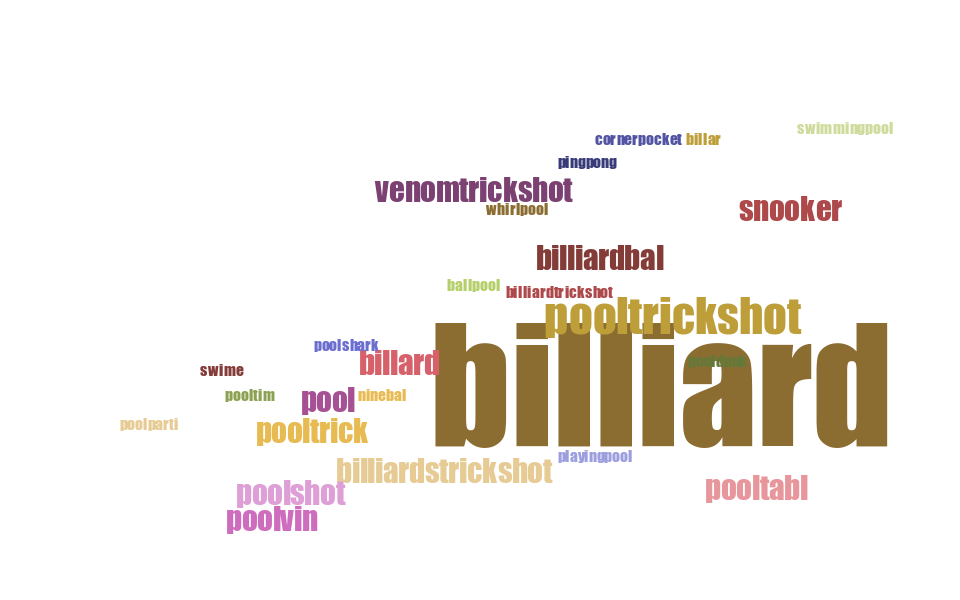}
 \includegraphics[width=0.23\textwidth]{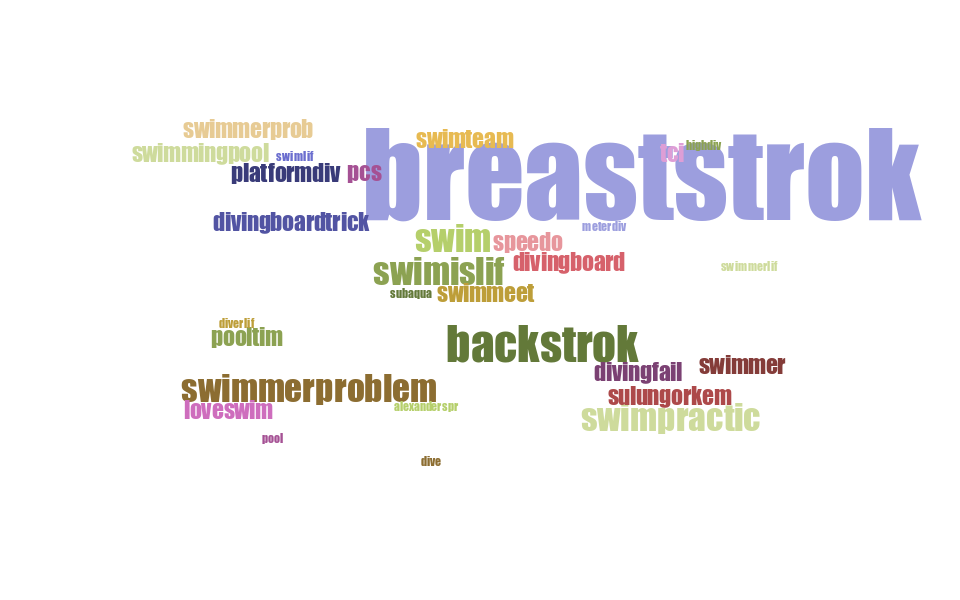}
 \caption{Illustration of effectiveness of Tag2Vec representation. 
 The word clouds represent 29 nearest neighbours in Tag2Vec space for
 the following queries : `basketball', `billiards', `pushup',
`baseball', `kayaking', and `breast stroke'. The word sizes are proportional
to similarity (inversely proportional of $L_2$ distances), query word being
the largest due to 100\% self-similarity. Note that tag words are stemmed, not spelled incorrectly.}
 \label{fig:tagNearby}
\end{figure}

\begin{figure*}[t!]
\captionsetup[subfigure]{labelformat=empty}
  \centering
  \subfloat[]{\raisebox{1.75in}{\rotatebox[origin=t]{90}{(a) Training Phase}}} 
  \subfloat[]{\includegraphics[width=0.88\textwidth]{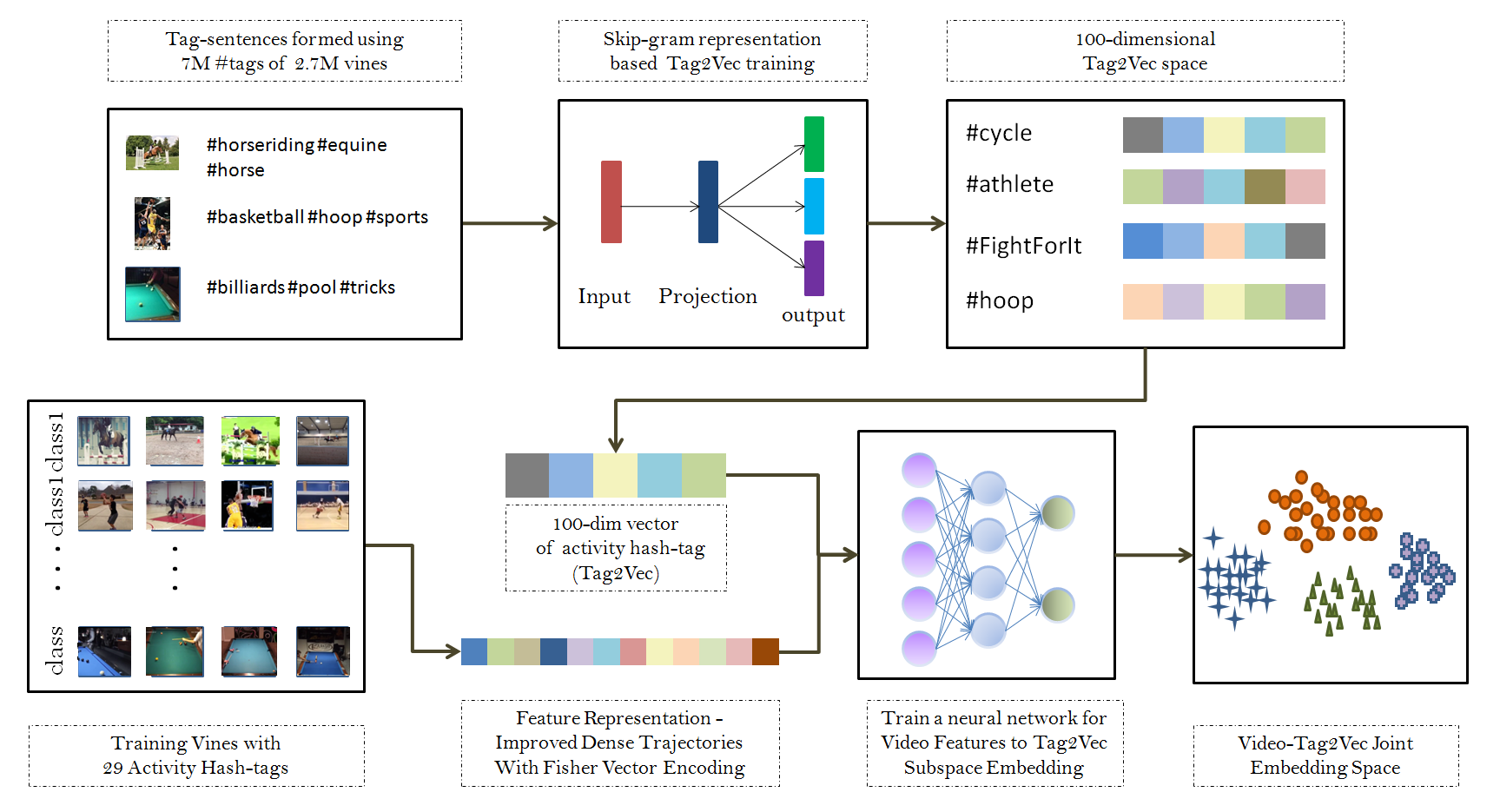}}\\
  \vspace{-5mm}
  \subfloat[]{\raisebox{0.75in}{\rotatebox[origin=t]{90}{(a) Testing Phase}}} 
  \subfloat[]{\includegraphics[width=0.88\textwidth]{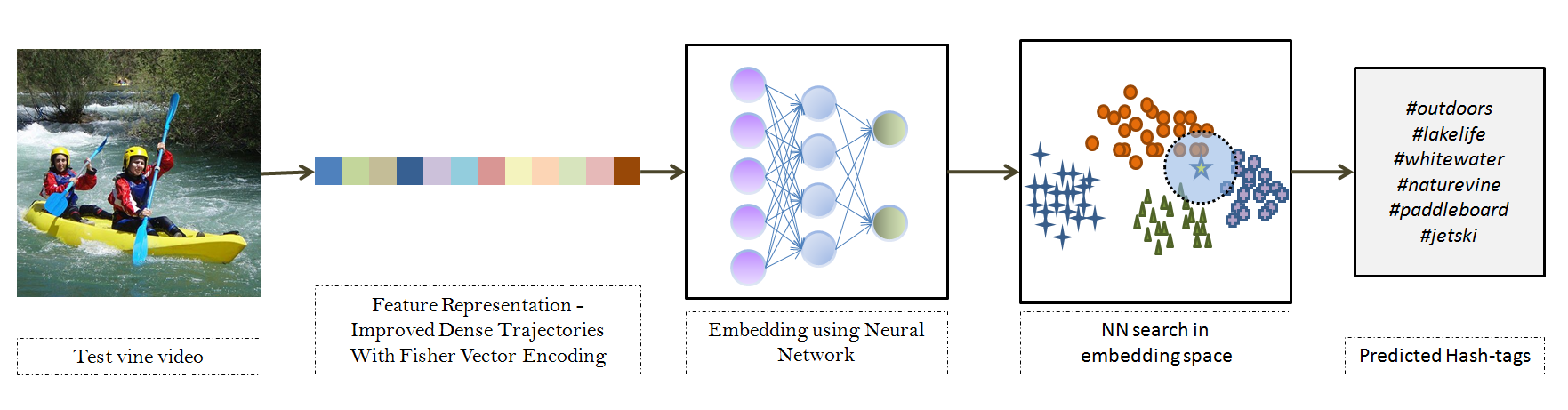}}
  \vspace{-5mm}
  \caption{Pictorial representation of our tag recommendation system} 
  \label{fig:flowdiagram}
\end{figure*}


%

The proposed system is trained using a large number of videos and associated hash-tags scraped from social 
media platform \url{vine.co}. Videos shared on this platform (commonly known as vines) 
are six seconds long, often captured by hand-held or wearable devices, with cuts and edits, and present a significantly 
wilder and more challenging distribution than traditional videos. For each uploaded video, the original poster also provides
hash-tags. Unlike tag words typically used as meta-data, hash-tags serve more of a social purpose to improve content
visibility and to associate content with social trends. Many hash-tags do not adhere to the commonly understood semantics of the natural language.
Due to this reason, we learn a new tag space representation \emph{Tag2Vec} instead of directly using 
semantically structured Word2Vec space of \cite{word2vec}. \autoref{fig:vinetagsnadshots} shows 
examples of a few vine videos and associated hash-tags. It can be observed that 
{\tt\#FitFluential} and {\tt\#Mr315} are non-word hash-tags but understandable social media jargons given the content.

As mentioned previously, our end-to-end training for \emph{video-to-tag} mapping consists of two stages, of learning the 
Tag2Vec representation, and of learning the video features to tag vector space embedding. Finally, given a query video, 
the learned embedding projects it to the tag space and nearby hash-tags are retrieved as suggestion. This process 
is outlined in \autoref{fig:flowdiagram}. In the following subsections, we explain (i) the hash-tag data and learning of 
Tag2Vec space, (ii) the video data and learning of visual to Tag2Vec space embedding, and finally (iii) retrieval for tag recommendation. 

\subsection{Hash-tag Data and Pre-processing}
To gather hash-tags data, we first use the 17,000 most common English words (as determined by n-gram frequency
analysis of the Google's Trillion Word Corpus \footnote{https://github.com/first20hours/google-10000-english}) as queries and 
retrieve a total of about $2.7$ million videos ($\sim$ 150 videos per query). 
We scrape the hash-tags corresponding to each retrieved video, remove all special characters, and perform \emph{stemming} on the hash-tags. 

Stemming is a popular technique in natural language processing (NLP) community for 
reducing words to a root form such that multiple inflections of a word reduce to 
the same root e.g. `fish', `fished', `fishing', and `fish-like' reduce to the stem 
`fish'. The stemmed hash-tag words for each video 
form a \emph{hash-tag sentence} leading to a total of $2.7$ million sentences. 
These $2.7$ million hash-tag sentences together form a text-corpus that we use 
to learn the Tag2Vec representation. 

\subsection{Learning Tag2Vec Representation}
\citet{word2vec} proposed efficient unsupervised neural network based 
methods to learn embeddings of semantic words in vector space using a large 
corpus of text data (web-based) consisting of 1.6 billion words. These methods either 
use continuous Bag of Words representation or skip-gram representation of text sequences to 
train a two-layer neural network. The resulting word embedding assigns every word in the corpus
to a vector in a 300-dimensional space. This word embedding is known as Word2Vec and the final vector
space is popularly referred to as Word2Vec space. The main advantage 
of this representation is that vector operations can be performed on words. 
This property is extremely useful, e.g. to compute similarity between 
words using vector space distances. 

Similar to this, we also train a neural network to learn hash-tag embeddings in 
a vector space and call it \emph{Tag2Vec}. Since, we work with much smaller 
data ($\sim$ 7 million unique tags and 2.7 million sentences), we use skip-gram 
representation which is more effective on small data and we also restrict the resulting 
space to be 100-dimensional. We use publicly available code\footnote{\url{https://code.google.com/p/word2vec/}} 
for learning the Tag2Vec embeddings. The training converges quickly 
($\sim$ 10 minutes) as we have a relatively small corpus of hash-tags. 
The resulting vector space enables us to perform 
vector operations on hash-tags. \autoref{fig:tagNearby} shows a 
word cloud representation of $30$ nearest neighbours in Tag2Vec space for query 
vectors corresponding to stemmed tag words, `basketball', `billiards', `pushup', 
`baseball', `kayaking', and `breast stroke'. The word sizes are inversely related to the 
distance from the respectively query words, the closer the words in vector space, the larger 
the font size. It can be observed that `basketball' tag has many tags 
with high similarity whereas `billiards' has fewer. It is also worth noting 
how contextual similarity is also well captured, for example `kayaking' tag has 
strong neighbours such as `state park' and `summer adventure'. 

To demonstrate that the learned Tag2Vec space is different from 
Word2Vec space, we list the top-10 near-neighbours for four 
action word queries in both spaces (see \autoref{tab:t2v_w2v}). 
It can be clearly seen that the tags retrieved using Tag2Vec 
space are more diverse and socially relevant. See particularly 
the results for `Polevault' and `Basketball' queries. 
We also measure similarity between pairs of tag vectors 
and see that the Tag2Vec space models social jargons well. 
For example, we noticed that tags like \#lol and \#laugh
have high similarity, \#lol is also has high similarity with 
\#fail owing to users tagging funny videos showing people 
failing at doing something, \#fight is closer to both \#win 
and \#fail. This shows that our tag2vec space is able to capture 
meaningful tag relationships and similarity.\\

\newcommand{\specialcellcen}[2][c]{%
          \begin{tabular}[#1]{@{}c@{}}#2\end{tabular}}   
          
\renewcommand{\arraystretch}{1.2}          
\begin{table*}
\resizebox{\linewidth}{!}{%
\centering
\begin{tabular}{ c  c c c c c c c c c c c }
\toprule
\specialcellcen{Query \\ Words} & \specialcellcen{Vector \\ Space} & \multicolumn{10}{c}{top-10 retrieval results} \\
\midrule
\multirow{2}{*}{{\bf Pushups}} & {\tt Word2Vec} & jumping jacks & pushup & situps & calisthenics & \specialcellcen{abdominal\\crunches} & \specialcellcen{pushups\\situps} & burpees & pullups & \specialcellcen{ab crunches} & \specialcellcen{squat\\thrusts}\\
\cmidrule(lr){2-12}
& {\tt Tag2Vec} & workout & gym & burpees & gymflow & gymday & exercise & muscleup & calisthenics & superset & gymrat\\
\midrule
\midrule
\multirow{2}{*}{{\bf Polevault}} & {\tt Word2Vec}& \specialcellcen{Ivan\\Ukhov} & \specialcellcen{Gulfiya\\Khanafeyeva} & 
\specialcellcen{Yaroslav\\Rybakov} & \specialcellcen{Tatyana\\Lysenko} & \specialcellcen{Anna\\Chicherova} & 
\specialcellcen{Andrey\\Silnov} & \specialcellcen{champion\\Tatyana} & \specialcellcen{Croatia\\Blanka Vlasic} 
& \specialcellcen{Svetlana\\Feofanova} & \specialcellcen{Olga\\Kuzenkova} \\
\cmidrule(lr){2-12}
& {\tt Tag2Vec}& USATF & athlete & tracknation & trackandfield & discusthrow & tooathlete & highjump & maxvelocity & blockstart & longjump \\
\midrule
\midrule
\multirow{2}{*}{{\bf Kayaking}} & {\tt Word2Vec} & canoeing & Kayaking & kayak & paddling & sea kayaking & rafting & \specialcellcen{whitewater\\kayaking} & \specialcellcen{kayaking\\canoeing} & \specialcellcen{rafting\\kayaking} & canoing\\
\cmidrule(lr){2-12}
& {\tt Tag2Vec} & statepark & lake & whitewater & paddleboard & outdoorsfinland & lagoon & lakelife & outdooraction & emeraldbay & boat\\
\midrule
\midrule
\multirow{2}{*}{{\bf Basketball}} & {\tt Word2Vec} & baskeball & volleyball & basketbal & basektball & hoops & soccer & softball & football & bas ketball & roundball\\
\cmidrule(lr){2-12}
& {\tt Tag2Vec} & dunk & ballislife & hoopmixtape & basketballvine & NBA & basketballneverstops & streetball & basketballhighlight & bball & dunkcity\\
\bottomrule
\end{tabular}
}%
\caption{Comparision of Tag2Vec and Word2Vec spaces. Top-10 nearest neighbour results shown for four query words. It can be seen that the tag words retrieved from 
Tag2Vec space are more diverse and socially relevant.}

\label{tab:t2v_w2v}
\end{table*}

\begin{figure*}[thb!]
  \centering
  \vspace{-2mm}
  \subfloat[Before embedding]{\includegraphics[width=0.35\textwidth]{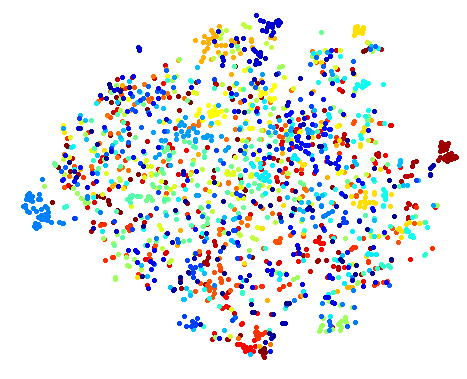}\label{fig:f1}}\quad
  \subfloat[After embedding]{\includegraphics[width=0.35\textwidth]{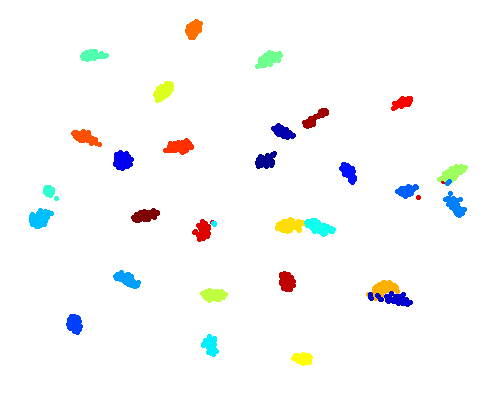}\label{fig:f2}}
  \caption{t-SNE visualization of the video features in fisher vector space and Tag2Vec space (after embedding)}
  \vspace{-2mm}
  \label{fig:beforeafterv2t}
\end{figure*}

\begin{table*}
\centering
\resizebox{\textwidth}{!}{%
 \begin{tabular}{ c c c c c c c c c c}
 \toprule
  Baseball & Basketball & Benchpress & Biking & Billiards &
   Boxing & Breaststroke & Diving & Drumming & Fencing \\
   Golf & HighJump & Horseriding & HulaHoop &  Juggling &
   Kayaking & Lunges & Nunchucks & Piano & PoleVault \\
    Pushups &  Yoyo & Salsa  & Skateboarding & Skiing  &
 Soccer & Swing & Tennis & Volleyball &  \\
 \bottomrule
 \end{tabular}
 }
 \caption{29 video categories used for training}
 \label{tab:categories}
\end{table*}

\subsection{Video Data and Feature Representation}

Our video data consists of vine video clips (vines) spanning 
29 categories. These categories are listed in \autoref{tab:categories}. 
As mentioned before, vines are short but wild and complex amateur video clips 
with heavy camera motion, cuts, and edits. Hence, they have high intra-class variance 
and direct similarity based approaches don't work well. We obtain 3000 useful videos with 
hash-tags after removing duplicates. We split these into training and testing sets of 
2740 and 260 videos respectively. For both sets of videos, we compute visual and 
motion features. 

In particular, we compute Improved Dense Trajectory (IDT) \cite{ImpDenseTraj} 
features which consist of HoG (histogram of oriented gradients), HoF (histogram of optical flow), 
and MBH (motion boundary histograms) features. These low level features capture global scene, motion and rate of motion information respectively. We encode the low-level IDT features 
using $1003676$ dimensional fisher vectors for better generalization \cite{fv1,fv2}. Fisher encoding relies on gaussian mixture model 
(GMM) computed over a large vocabulary of low-level features. We use UCF51 action recognition dataset \cite{UCF101} 
to compute generalized vocabulary for GMM estimation. 
For IDT extraction, we use the code\footnote{\url{https://lear.inrialpes.fr/people/wang/improved_trajectories}} made available by the 
authors. For computing GMM parameters and fisher vectors, we use the VLFeat Computer Vision 
library \cite{vlfeat}.

\subsection{Learning Video to Hash-tag Embedding}
In the first step, a 100-dimensional vector space, representative of the hash-tag distribution 
has been learned. Next we need to learn a mapping function that can project the 
$1003676$ dimensional video features (fisher encoded IDTs) to the $100$ dimensional tag vector space.

\citet{zslearn} proposed a method to learn a mapping from visual features (images) to word vectors (word2vec space) for detecting objects 
in a cross-modal zero-shot framework. We adopt this cross-modal learning approach to learn a mapping from fisher vectors to tag vectors. 
Similar to \cite{zslearn}, we train a neural network with \texttt{(fisher vector, tag word)} pairs for each of the 2770 training videos 
to learn a non-linear embedding function from video features too Tag2Vec space. The tag word is the same as the category/class label of 
the training video. 
For learning this embedding function, we use the publicly available code for zero-shot learning \footnote{\url{https://code.google.com/p/word2vec/}}. 
The neural network is set up to have $600$ hidden nodes and maximum iterations are set to $1000$ as we have more categories. 
Training this network with our data took approximately 2 hours. 

\autoref{fig:beforeafterv2t} shows the t-SNE (t-Stochastic Neighborhood Embedding) visualization of training features in fisher
vector space and Tag2Vec space. It can be clearly seen that after embedding the the training vectors form distinct clusters
around their category words. Once the embedding function is learned, a query video (belonging to these 29 categories) 
can be directly mapped to the tag space and relevant hash-tags can be recommended. 
In the next section, we explain the hash-tag recommendation mechanism.

\subsection{Tag Suggestion Metrics}
Given a query video, we first compute its fisher vector representation. 
We then use the learned embedding function to project the query 
fisher vector in the learned Tag2Vec space. We utilize 
a simple nearest neighbour approach based on $L_2$ distance 
to retrieve potentially relevant 
hash-tags for a given query video. It can be seen that we do not 
directly compare the test/query vector to any of the training video 
vectors, neither in fisher vector space, not after the embedding. 
This is advantageous in terms of retrieval time and memory 
because, (i) there is no need to store the training set 
but only the Tag2Vec model and the fisher vectors to Tag2Vec space embedding function; 
and (ii) redundant comparisons are avoided. The tag words 
retrieved from the Tag2Vec space are stem words. Since we 
cannot suggest stems as hash-tags we need to convert a stemmed tag 
to its proper form. However, each stem 
corresponds to multiple tags, e.g. `beauty', `beautiful', 
`beautifully' would all map to stem word `beauti'. In our system, 
for a particular stemmed tag, we pick the most commonly used word 
in our hash-tag corpus from among all corresponding inflections of that 
stem. This de-stemming approach is a bit limiting as many 
variations of the same stem words would always be rejected. 
A better approach based on parts of speech (PoS) tagging and 
edit distance can replace this. 
The time complexity of tag suggestion depends on the 
dimensionality of the tag space ($d$) and number of tags ($N$), 
$O(N*d)$. Our MATLAB implementation takes around 1 second 
for video-to-tag space embedding and less than a second for 
near-neighbour tag retrieval. 
For large-scale application, the simple nearest neighbour approach 
can be replaced by better retrieval mechanisms for efficiency and 
robustness. 

\section{Experiment \& Results}
\label{sec:exp}

\begin{figure}[t!]
  \centering
  \includegraphics[width=0.45\textwidth]{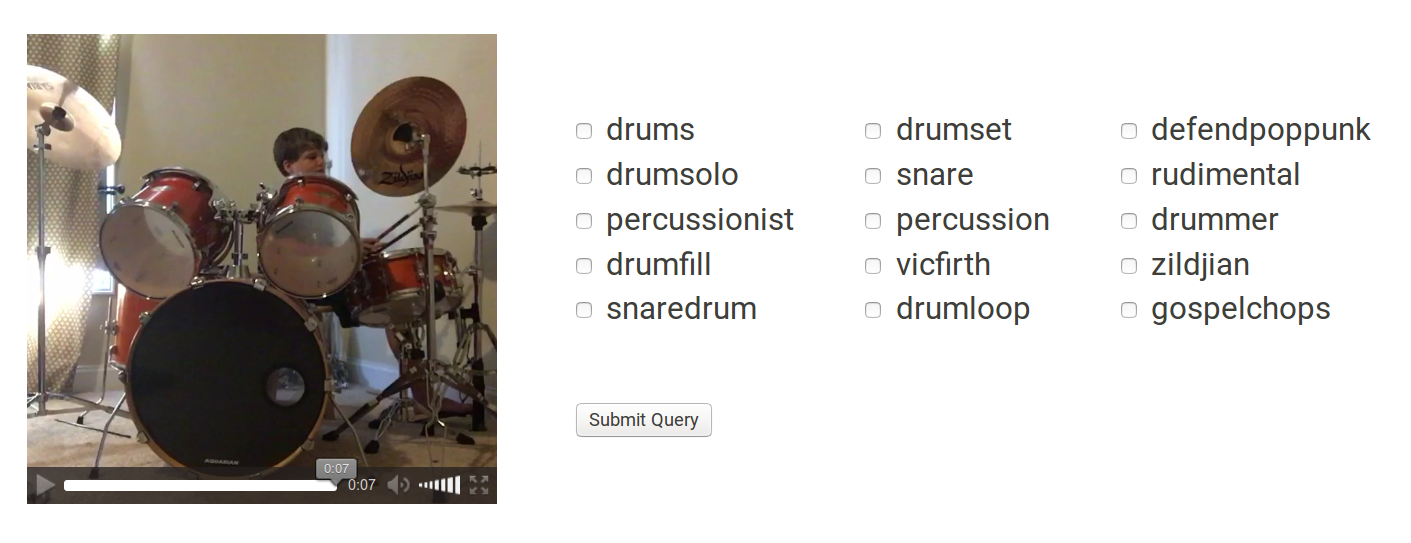}
  \caption{A screen shot of our user annotation (survey) platform.}
  \label{fig:userstudy}
\end{figure}

We conduct experiments to both qualitatively and quantitatively validate the tags suggested by our approach. 
Users who are familiar with social networking jargons are asked to take a survey.
In user survey, each user is shown a set of vines with 15 recommended hashtags per vine.
Users can pause/replay the vine and select the hashtags they consider can be used with the 
shown vine. \autoref{fig:userstudy} shows a snapshot of the user study session. 
We perform these experiment with $14$ users where every user marks each vine once.
The testing set contains $270$ vines with approximately $9$ vines per class. 
Vines are wild and intra-class variations are quite drastic. Hence, we compute the 
average relevance scores per class for better understanding the cases where the 
system performs better or worse. As our dataset consists of wild, unconstrained, 
and unfiltered vines, there are cases where users don't find any suggested hashtags as 
relevant. To see whether this happens for specific classes or it is uniform across classes, 
we also collect the data for the number of vines per class for which users didn't find 
any recommended tag as relevant. 

\begin{figure*}[t!]
 \centering
 \begin{tabular}{c  c  c}
 \includegraphics[width=0.3\linewidth]{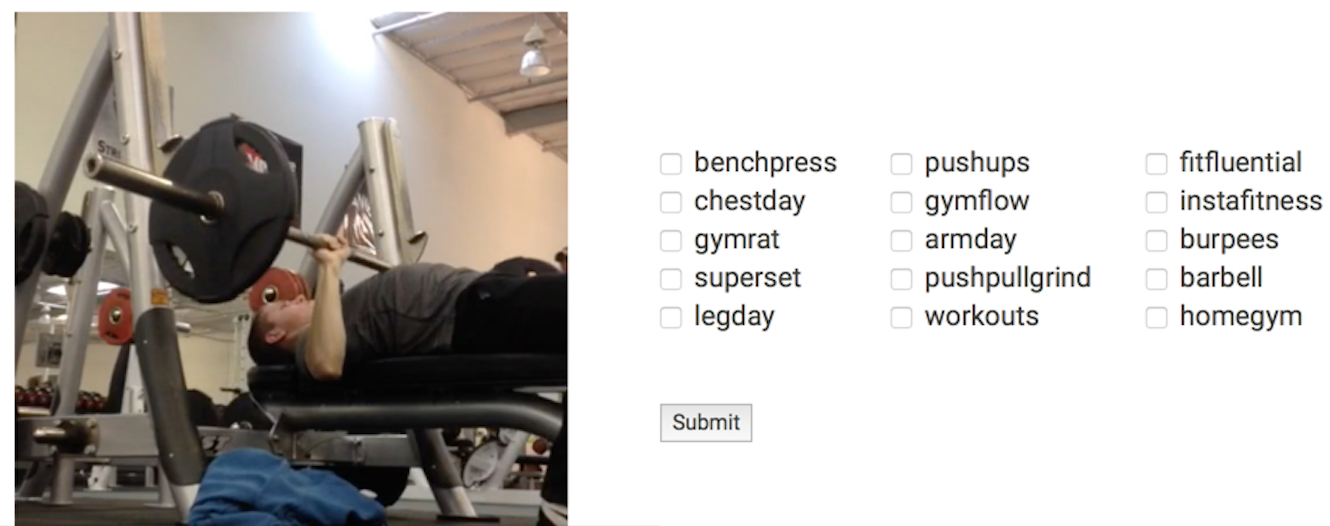} & 
 \includegraphics[width=0.3\linewidth]{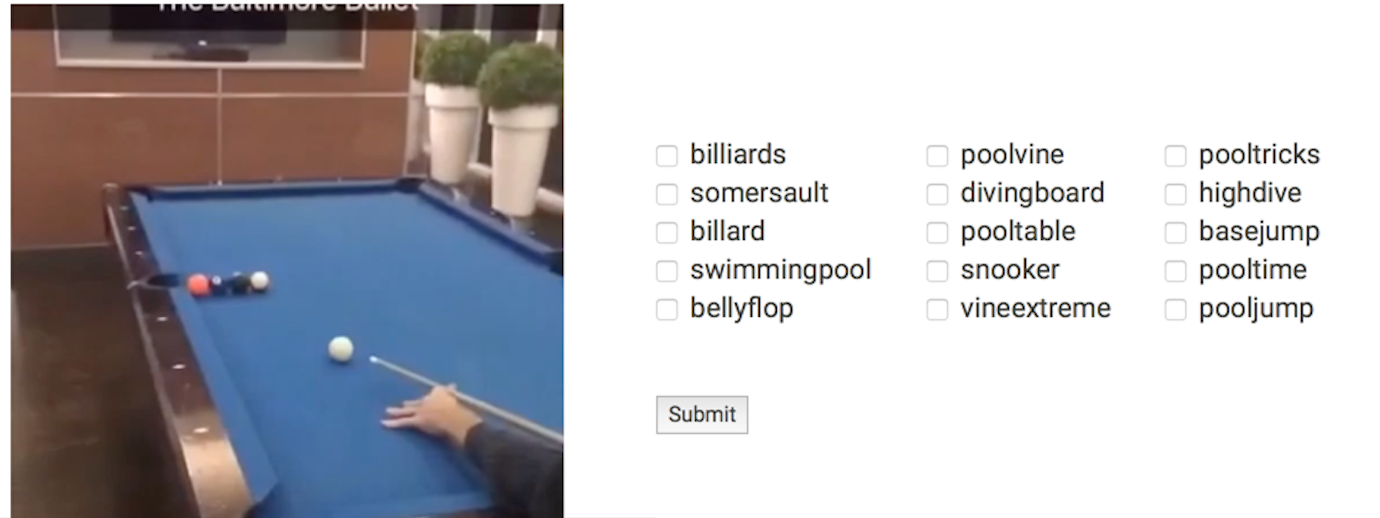} &
 \includegraphics[width=0.3\linewidth]{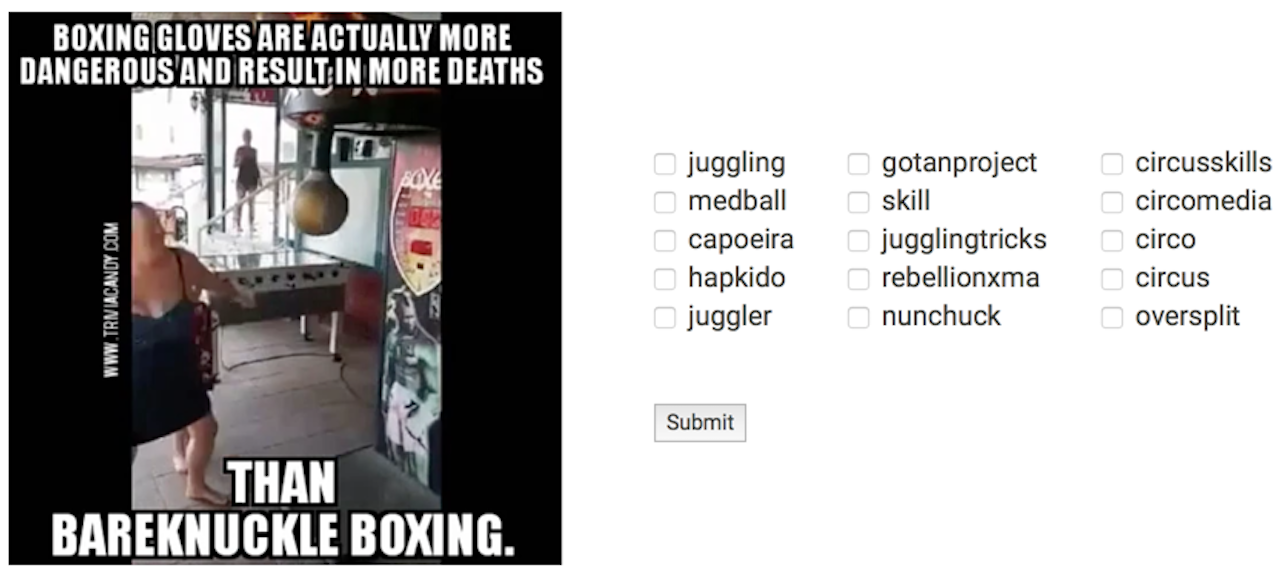} \\ \\
 \includegraphics[width=0.3\linewidth]{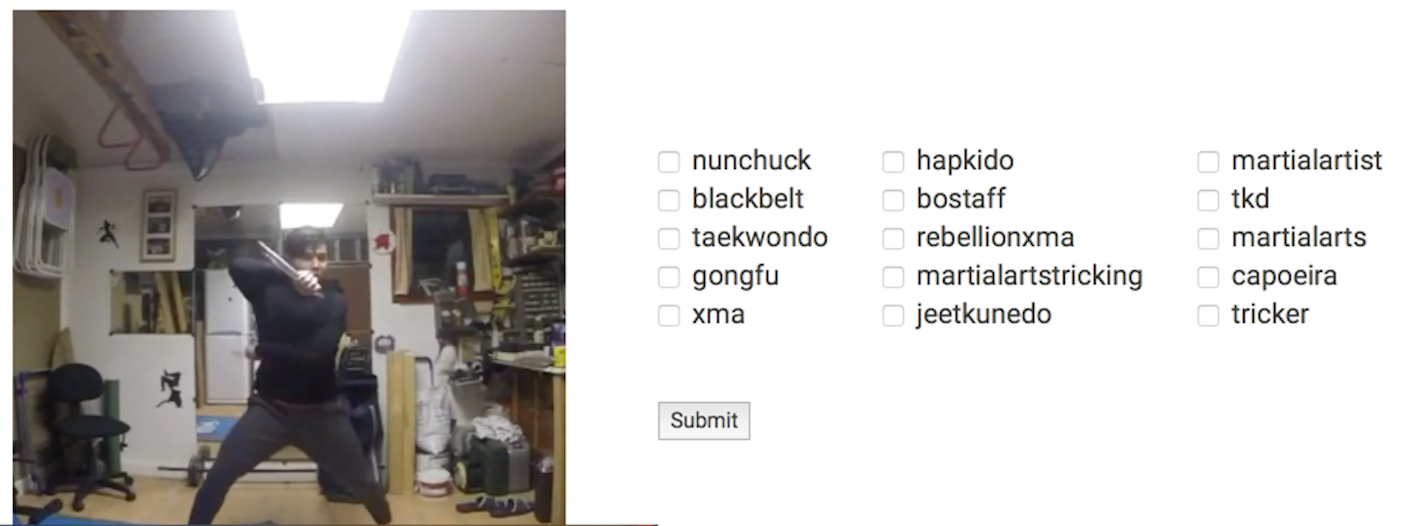} &
 \includegraphics[width=0.3\linewidth]{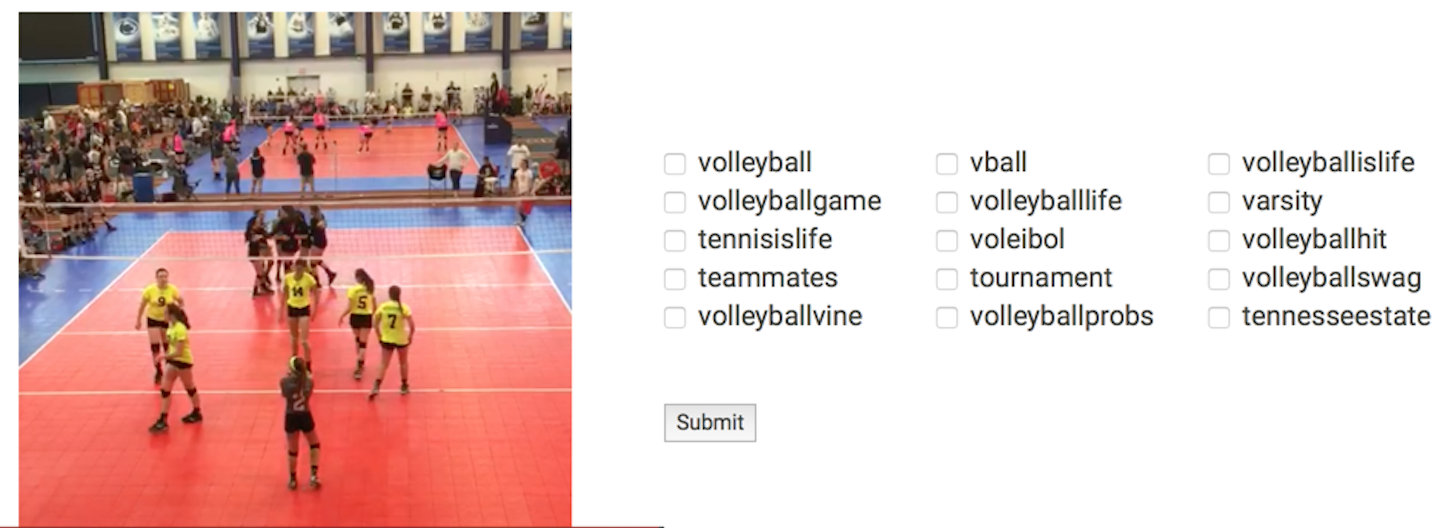} &
 \includegraphics[width=0.3\linewidth]{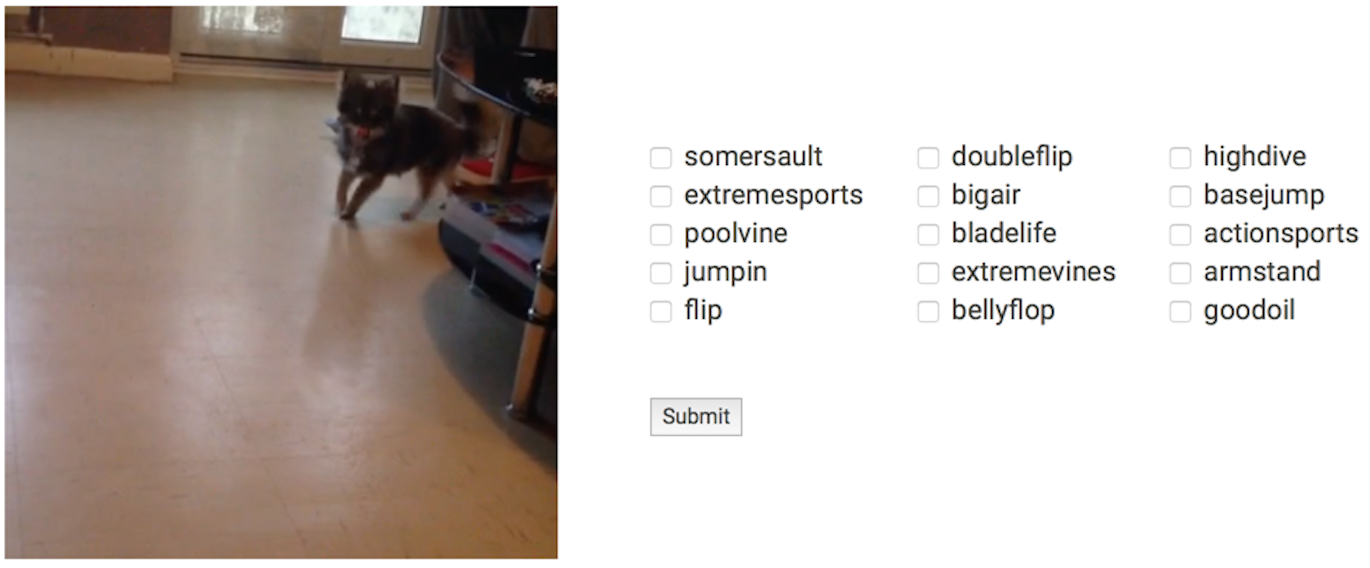} 
 \end{tabular}
 \caption{Hash tags suggested by our framework for the given video clip. First two columns show relevant hash-tag suggestions as predicted 
 by our proposed model (like armday, benchpress, instafitness etc. for top left image). Last column shows two failure cases. }
\label{fig:vidResults}
\end{figure*}
\begin{figure*}[t!]
\centering
 \includegraphics[width=0.48\linewidth]{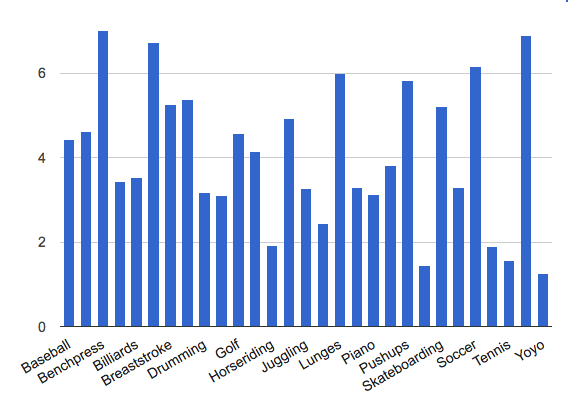} \,
 \includegraphics[width=0.48\linewidth]{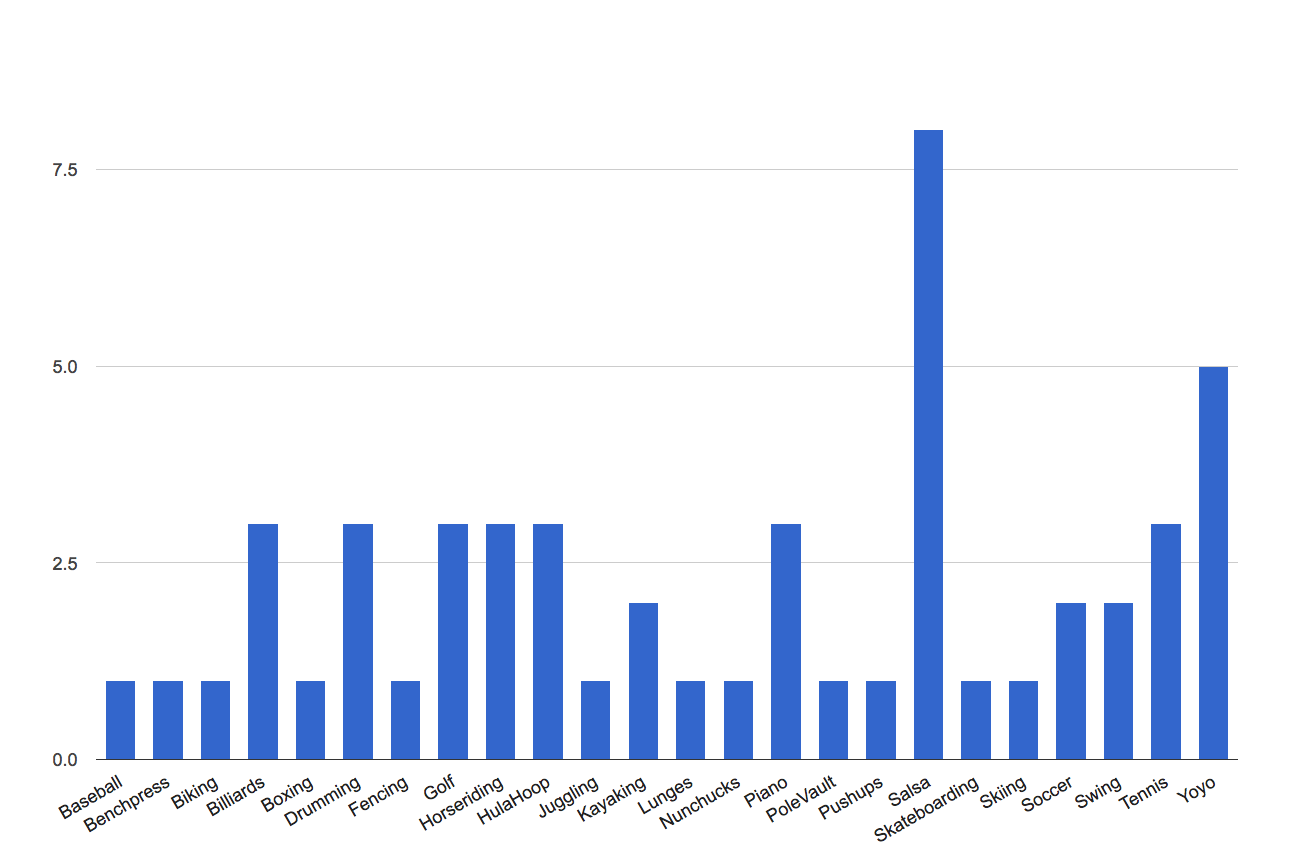}
 \caption{Left image shows the average number of relevant tags marked by the users for each class out of 15 suggest tags. 
 Right image shows average number of videos across all users per class for which there were no relevant tags found out of entire test dataset of
 ~ 50 videos per class}
 \label{fig:averagescore}
\end{figure*}


The plot in \autoref{fig:averagescore} (left) shows the average number of relevant tags suggested for each class.
The plot in \autoref{fig:averagescore} (right) shows the class-wise distribution of vines with no relevant hashtags. 
Benchpress contains the highest number of tags, $7$, on an average followed by Volleyball at $6.88$.
Yoyo \& Salsa are the worst performers with $1.25$ \& $1.45$ tags respectively on an average. 
In total $52$ vines out of the $270$ didn't contain a single relevant hashtag. 
Upon viewing these vines, we noticed that these vines in majority were the ones 
which visually and textually contained no information pertaining to the action tag.
For example, a person talking about how good boxing is might contain relevant hashtags as 
assigned by the uploader but as we don't process auditory modality and training of the 
embedding function relies only on visual features, the system is unavailable to correctly 
map such vines to the relevant concepts. \autoref{fig:vidResults} shows some 
examples of success and failure cases for qualitative evaluation. 

The overall number of relevant tags suggested for a vine is $4.03$ out of $15$ which 
is $27\%$ of the tags suggested for each vine. Based on the hash-tag statistics collected 
by scraping the vine platform, we observed that $4.79$ is the average number of hashtags 
associated with a typical vine (based on 2.5 million entries). One thing to note is that 
not all the tags in the ground-truth data are relevant hence this number is likely 
to come down. By suggesting $15$ tags we are able to reproduce a similar number where an 
uploader finds $4$ tags relevant to the vine which suggests that our system performs well 
even for such unconstrained videos.

\section{Conclusion and Future Works}
\label{sec:conclusion}
In summary, we present a method to automatically suggest hash-tags for short social video 
clips. Hash-tags are noisy and have ambiguous semantics. We learn a vector space which
we show is able to capture these semantics. We call this vector space Tag2Vec.
Also for automatically annotating hash-tags for a given video we learn a neural network based
embedding function. We work on a self gathered dataset of wild short video clips of 29 categories. 
The embedding function embeds any query video to our Tag2Vec space from which we propose 
hash-tags using simple nearest neighbour retrieval. We show that our Tag2Vec space
has desired semantic structure and we are able to suggest relevant hash-tags for the query videos.
In future, we would like to explore sentimental vs.\ contextual relevance in the Tag2Vec space and 
would also like to incorporate relevance metric that depends on the evolving popularity and 
trends of the hash-tags. 

\section{Acknowledgement}
We thank Shivam Kakkar and Siddhartha Gairola for their 
crucial contributions in developing the tool for user study. 
Rajvi Shah and Saurabh Saini are supported respectively 
by Google and TCS PhD fellowships. We thank these 
organizations for their valuable support. 

{
  \bibliographystyle{plainnat}
  \bibliography{main.bib}
}

\end{document}